\documentclass{article}

\usepackage{arxiv}

\usepackage[utf8]{inputenc} 
\usepackage[T1]{fontenc}    
\usepackage{hyperref}       
\usepackage{url}            
\usepackage{booktabs}       
\usepackage{amsfonts}       
\usepackage{nicefrac}       
\usepackage{microtype}      
\usepackage{lipsum}
\usepackage{graphicx}
\graphicspath{ {./images/} }
\usepackage{lscape}

\title{Analysis and prediction of heart stroke from ejection fraction and serum creatinine using LSTM deep learning approach}

\author{
 Md Ershadul Haque \\
 School of Computing, Mathematics and Engineering\\
  Charles Sturt University\\
  Bathurst, 2795 \\
  Australia\\
  \texttt{mhaque@csu.edu.au} \\
   \And
  Salah Uddin \\
 Dept. of Electrical and Electronic Engineering\\
 Feni University\\
 Feni,3900 \\
  Bangladesh\\
  \texttt{salahuddin@feniuniversity.edu.bd} \\
  \And
  Md Ariful Islam \\
 Dept. of Electrical and Electronic Engineering\\
 Feni University\\
 Feni,3900 \\
  Bangladesh\\
  \texttt{eng.arif968@gmail.com} \\
  \And
  Amira Khanom \\
 Dept. of Electrical and Electronic Engineering\\
 Z.H. Sikder University of Science and Technology\\
 Sariatpur \\
  Bangladesh\\
  \texttt{amirakhanom5@gmail.com} \\
  \And
  Abdulla Suman\\
  Department of Clinical Medicine\\
  Macquarie University\\
  Sydney \\
  Australia\\
  abdulla.suman@mq.edu.au \\
   \And
  Manoranjan Paul \\
  School of Computing, Mathematics and Engineering\\
  Charles Sturt University\\
  Bathurst, 2795 \\
  Australia\\
  \texttt{mpaul@csu.edu.au} \\
 \\
\\
}

\begin{document}
\maketitle
\begin{abstract}
The combination of big data and deep learning is a world-shattering technology that can greatly impact any objective if used properly. With the availability of a large volume of health care datasets and progressions in deep learning techniques, systems are now well equipped to predict the future trend of any health problems. From the literature survey, we found the SVM was used to predict the heart failure rate without relating objective factors. Utilizing the intensity of important historical information in electronic health records (EHR), we have built a smart and predictive model utilizing long short-term memory (LSTM) and predict the future trend of heart failure based on that health record. Hence the fundamental commitment of this work is to predict the failure of the heart using an LSTM based on the patient's electronic medicinal information. We have analyzed a dataset containing the medical records of 299 heart failure patients collected at the Faisalabad Institute of Cardiology and the Allied Hospital in Faisalabad (Punjab, Pakistan). The patients consisted of 105 women and 194 men and their ages ranged from 40 and 95 years old. The dataset contains 13 features, which report clinical, body, and lifestyle information responsible for heart failure. We have found an increasing trend in our analysis which will contribute to advancing the knowledge in the field of heart stroke prediction. 
 
\end{abstract}


\section{Introduction}
\begin{figure}[h]
    \centering
    \includegraphics[width=0.5\textwidth,height=4cm]{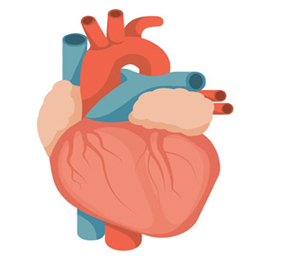}
    \caption{schematic illustration of the anatomy of heart.}
    \label{f1}
\end{figure}

The heart is a mid line, valvular, muscular pump that is cone-shaped and the size of a fist. In adults, it weighs 300 grams and lies in the middle mediating of the thorax. The inferior (diaphragmatic) surface sits on the central tendon of the diaphragm, whereas the base faces posteriorly and lies immediately anterior to the esophagus and (posterior to that) the descending aorta. The base comprises mainly the left atrium. The left surface (left ventricle) and right surface (right atrium) are each related laterally to a lung and a phrenic nerve in the fibrous pericardium. Although essentially a mid line structure, one-third of the heart lies to the right of the mid line and two-thirds to the left.
The interventricular septum bulges to the right because the wall of the left ventricle is much thicker (10 mm) than that of the right ventricle (3-5 mm). It also lies obliquely across the heart, almost in the coronal plane, such that the anterior surface of the heart is two-thirds right ventricle and one-third left ventricle; the proportions are reversed on the inferior surface. The thicker, muscular part of the interventricular septum is formed from the ventricular walls. The muscles of the four chambers and the four valves are attached to, and supported by, a figure-of-eight-shaped fibrous skeleton comprising a central fibrous body and extensions (fila coronaria) that surround the valves. This skeleton both divides and separates the atria electrically from the ventricles and is the remnant of the atrioventricular (AV) cushions. The thinner membranous part of the interventricular septum is formed from the lowest aspect of the spiral valve (neural crest cells), which divides the truncus arteriosus into the aorta and pulmonary trunk\cite{c1}.
The pericardium holds and protects the heart, but provides sufficient potential space for filling and emptying of the chambers. The outer layer is the tough fibrous pericardium, which blends with the adventitia of the aorta, the pulmonary trunk, the superior vena cava and the central tendon of the diaphragm. Within this, there are two layers of serous pericardium:
\begin{itemize}
\item a visceral layer, surrounding the heart
\item a parietal layer, lining the inner surface of the fibrous pericardium.
\end{itemize}
These two layers of serous pericardium are continuous with each other as they reflect off the major vessels behind and above the heart. Figure \ref{f1} shows the schematic illustration of the human heart. The reflection, posteriorly, between the pulmonary veins is termed the ‘oblique sinus’ of the pericardium. The plane between the superior vena cava and the pulmonary veins posteriorly, and the aorta and pulmonary trunk anteriorly, made by the folding of the heart, is termed the ‘transverse sinus’ of the pericardium. The visceral layer and the heart itself are supplied by sympathetic nerves from the cardiac plexuses; these in turn carry general visceral afferent fibres to the vertebral levels from which the sympathetic supply arises, which are the three cervical sympathetic ganglia and the T1-5 ganglia e this explains why cardiac pain is referred to the neck, chest and arm\cite{c2}.
Heart failure is a syndrome with symptoms and signs caused by cardiac dysfunction, resulting inreduced longevity. To establish a diagnosis of heart failure, the European Society of Cardiology guidelines warrant the presence of symptoms and signs, objective evidence of cardiacdys function (preferably by echocardiography), and, in case of remaining doubt, a favourable response to treatment directed towards heart failure. To support the failing heart numerouscompensatory mechanisms occur, including activation of the neurohormonal system \cite{c3}. An increase in natriuretic peptide concentrations (particularly B type natriuretic peptide) is considered a hallmark of heart failure. The diagnosis of heart failure, especially when relying solely on symptoms and signs (which is often the case in primary care), is fraught with difficulties. Many patients deemed to have heart failure will simply be found to be obese, have a poor physical condition, pulmonary disease, orischaemia on further examination. Evidence is accumulating that normal natriuretic peptide levels and a normal electrocardiogram should lead to a reconsideration of a diagnosis of heart failure\cite{c4,c5,c6}

We focused mainly on some issues. A few of them are:
\begin{itemize}
    \item  As the day proceeds ahead, so the hospitalized peoples related to heart failure is increasing more and more. So, we need proper distribution of medical equipment throughout the country. Also the government need proper direction to allocate the budget for the increasing the medical facility. Hence a proper prediction of heart failure is necessary.
    \item  As the people are not getting much information about the number of heart failure case, they are being subconscious about the disease. This causes the increasing trends in heart failure. So a proper prediction may make people conscious about the severity of the heart failure.
    \item  A lot of factors are responsible for the heat failure. Sideline to the doctoral process, a numerical study is very necessary to identify the most influential factor for the heart failure.
\end{itemize}

The main aims of this work are to:
\begin{enumerate}
    \item  Predict the heart failure trends for the different aged people considering various factors responsible to the heart failure, such as Anaemia, Diabetes, Blood Pressure, Smoking etc. using Long Short Term Memory networks(LSTM) model.
    \item  Another objective of our research work is to increase the confidence of the LSTM model.
\end{enumerate}

\section{Related Work}
Predicting mortality is important in patients with heart failure (HF). However, current strategies for predicting risk are only modestly successful, likely because they are derived from statistical analysis methods that fail to capture prognostic information in large data sets containing multi-dimensional interactions.Article \cite{c7} used a machine learning algorithm to capture correlations between patient characteristics and mortality.
HF is a frequent health problem with high morbidity and mortality, increasing prevalence and escalating healthcare costs. By calculating a HF survival risk score based on patient-specific characteristics from Electronic Health Records (EHRs), authors of the paper \cite{c8} identify high-risk patients and apply individualized treatment and healthy living choices to potentially reduce their mortality risk.
Physicians classify patients into those with or without a specific disease.Authors compared the performance of these classification methods with those of conventional classification trees to classify patients with heart failure according to the following sub-types: heart failure with preserved ejection fraction (HFPEF) vs. heart failure with reduced ejection fraction (HFREF)\cite{c9}.
A study sought to develop models for predicting mortality and HF hospitalization for outpatients with HF with preserved ejection fraction (HFpEF) in the TOPCAT (Treatment of Preserved Cardiac Function Heart Failure with an Aldosterone Antagonist) trial\cite{c10}.
It also carries out an analysis including the follow-up month of each patient: even in this case, serum creatinine and ejection fraction are the most predictive clinical features of the data set, and are sufficient to predict patients’ survival \cite{c11}.
In the literature, a lot of studies and discussions on ML techniques i.e classification algorithms, regression algorithms, DNN, etc. that are used for HF are found. From the survey it is cleared that the LSTM method can be more important for the prediction of the HF \cite{c12}.

\begin{figure}[h]
    \centering
    \includegraphics[width=0.5\textwidth,height=4cm]{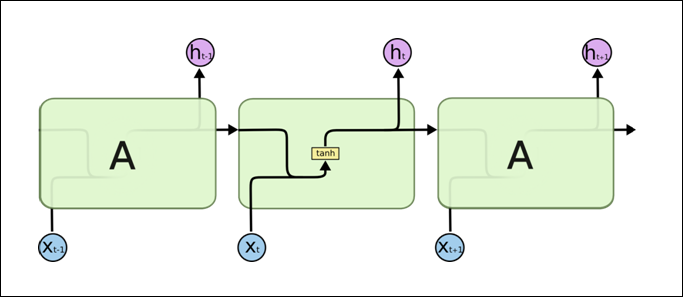}
    \caption{The repeating module in a standard RNN contains a single layer.}
    \label{f2}
\end{figure}

LSTM shown in Figure  \ref{f2} is a special kind of RNN , capable of learning long-term dependencies. It was introduced by\cite{c13}, and was refined and popularized by many people in following work. It works tremendously well on a large variety of problems, and is now widely used. LSTMs are explicitly designed to avoid the long-term dependency problem.  Information for long periods of time is practically their default behavior, not something they struggle to learn. All RNN has the form of a chain of repeating modules of neural network as shown. In standard RNN, this repeating module will have a very simple structure, such as a single tanh layer\cite{c14}.

\begin{figure}[h]
    \centering
    \includegraphics[width=0.5\textwidth,height=4cm]{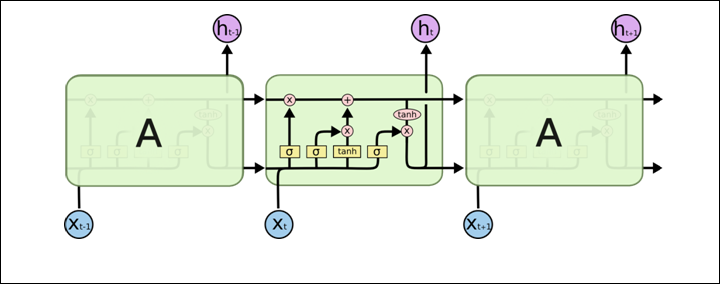}
    \caption{The repeating module in an LSTM contains four interacting layers.}
    \label{f3}
\end{figure}

LSTM also has this chain like structure, but the repeating module has a different structure.\cite{c15} Instead of having a single neural network layer, there are four, interacting in a very special way as shown in figure \ref{f3} . 

\begin{figure}[h]
    \centering
    \includegraphics[width=0.5\textwidth,height=4cm]{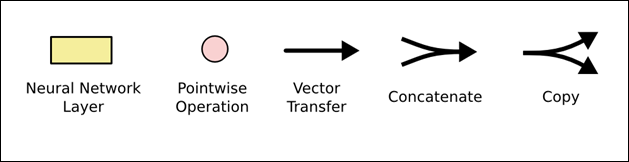}
    \caption{Notations used for describing the LSTM}
    \label{f4}
\end{figure}

Figure \ref{f4} illustrate the notation to get comfortable with the LSTM model. In the above diagram, each line carries an entire vector, from the output of one node to the inputs of others. The pink circles represent point wise operations, like vector addition, while the yellow boxes are learned neural network layers. Lines merging denote concatenation, while a line forking denote its content being copied and the copies going to different locations\cite{c16}.

The LSTM does have the ability to remove or add information to the cell state, carefully regulated by structures called gates. Gates are a way to optionally let information through. They are composed out of a sigmoid neural net layer and a point wise multiplication operation \cite{c17}.

\section{Data set}
 We analyzed a data set containing the medical records of 299 heart failure patients collected at the Faisalabad Institute of Cardiology and at the Allied Hospital in Faisalabad(Punjab, Pakistan), during April–December 2015.The patients consisted of 105 women and 194 men, and their ages range between 40 and 95 years old (Table 3.2)\cite{c18}. All 299 patients had left ventricular systolic dysfunction and had previous heart failures that put them in classes III or IV of New York Heart Association (NYHA) classification of the stages of heart failure. The data set contains 13 features, which report clinical, body, and lifestyle information (Table 3.1), that we briefly describe here. Some features are binary: anaemia, high blood pressure, diabetes, sex, and smoking. The hospital physician considered a patient having anaemia if haematocrit levels were lower than 36\%. Unfortunately, the original data set manuscript provides no definition of high blood pressure\cite{c19}. Regarding the features, the creatinine phosphokinase (CPK) states the level of the CPK enzyme in blood. When a muscle tissue gets damaged, CPK flows into the blood. Therefore, high levels of CPK in the blood of a patient might indicate a heart failure or injury. The ejection fraction states the percentage of how much blood the left ventricle pumps out with each contraction. The serum creatinine is a waste product generated by creatinine, when a muscle breaks down. Especially, doctors focus \cite{c20}.
 \begin{table}[]
\centering
\caption{A few rows of the data-sets}
\label{tab1_boldsymbol}
\resizebox{\textwidth}{!}{%
\begin{tabular}{|l|l|l|l|l|l|l|l|l|l|l|l|l|}
\hline
Age & Anemia & \begin{tabular}[c]{@{}l@{}}Creatinine Phosphoki-\\    \\ nase\end{tabular} & Diabetes & Ejection Fraction & \begin{tabular}[c]{@{}l@{}}High\\    \\ Blood\\    \\ Pressure\end{tabular} & Platelets & \begin{tabular}[c]{@{}l@{}}Serum\\    \\ Creatinine\end{tabular} & \begin{tabular}[c]{@{}l@{}}Serum\\    \\ Sodium\end{tabular} & Sex & Smoking & Time & Death event \\ \hline
75  & 0      & 582                                                                        & 0        & 20                & 1                                                                           & 265000    & 1.9                                                              & 130                                                          & 1   & 0       & 4    & 1           \\ \hline
55  & 0      & 7861                                                                       & 0        & 38                & 0                                                                           & 263358    & 1.1                                                              & 136                                                          & 1   & 0       & 6    & 1           \\ \hline
65  & 0      & 146                                                                        & 0        & 20                & 0                                                                           & 162000    & 1.3                                                              & 129                                                          & 1   & 1       & 7    & 1           \\ \hline
50  & 1      & 111                                                                        & 0        & 20                & 0                                                                           & 210000    & 1.9                                                              & 137                                                          & 1   & 0       & 7    & 1           \\ \hline
65  & 1      & 160                                                                        & 1        & 20                & 0                                                                           & 327000    & 2.7                                                              & 116                                                          & 0   & 0       & 8    & 1           \\ \hline
90  & 1      & 47                                                                         & 0        & 40                & 1                                                                           & 204000    & 2.1                                                              & 132                                                          & 1   & 1       & 8    & 1           \\ \hline
75  & 1      & 246                                                                        & 0        & 15                & 0                                                                           & 127000    & 1.2                                                              & 137                                                          & 1   & 0       & 10   & 1           \\ \hline
60  & 1      & 315                                                                        & 1        & 60                & 0                                                                           & 454000    & 1.1                                                              & 131                                                          & 1   & 1       & 10   & 1           \\ \hline
\end{tabular}%
}
\end{table}

\begin{table}[]
\centering
\caption{Meanings, measurement units, and intervals of each feature of the dataset}\label{tab1}
\resizebox{\textwidth}{!}{%
\begin{tabular}{|l|l|l|l|l|l|l|l|l|l|l|l|l|}
\hline
Feature                    & Explanation                                                 & Measurement      & Range                      & Ejection Fraction & \begin{tabular}[c]{@{}l@{}}High\\    \\ Blood\\    \\ Pressure\end{tabular} & Platelets & \begin{tabular}[c]{@{}l@{}}Serum\\    \\ Creatinine\end{tabular} & \begin{tabular}[c]{@{}l@{}}Serum\\    \\ Sodium\end{tabular} & Sex & Smoking & Time & Death event \\ \hline
Age                        & Age of the   patient                                        & Years            & {[}40, ..., 95{]}          & 20                & 1                                                                           & 265000    & 1.9                                                              & 130                                                          & 1   & 0       & 4    & 1           \\ \hline
Anaemia                    & Decrease of   red blood cells or hemoglobin                 & Boolean          & 0, 1                       & 38                & 0                                                                           & 263358    & 1.1                                                              & 136                                                          & 1   & 0       & 6    & 1           \\ \hline
High blood   pressure      & If a patient   has hypertension                             & Boolean          & 0,1                        & 20                & 0                                                                           & 162000    & 1.3                                                              & 129                                                          & 1   & 1       & 7    & 1           \\ \hline
Creatinine   phosphokinase & Level of the Creatinine   phosphokinase enzyme in the blood & mcg/L            & 23, ..., 7861{]}           & 20                & 0                                                                           & 210000    & 1.9                                                              & 137                                                          & 1   & 0       & 7    & 1           \\ \hline
Diabetes                   & If the patient   has diabetes                               & Boolean          & 0, 1                       & 20                & 0                                                                           & 327000    & 2.7                                                              & 116                                                          & 0   & 0       & 8    & 1           \\ \hline
Ejection   fraction        & Percentage of   blood leaving the heart at each contraction & Percentage       & {[}14, ..., 80{]}          & 40                & 1                                                                           & 204000    & 2.1                                                              & 132                                                          & 1   & 1       & 8    & 1           \\ \hline
Sex                        & Woman or man                                                & Binary           & 0, 1                       & 15                & 0                                                                           & 127000    & 1.2                                                              & 137                                                          & 1   & 0       & 10   & 1           \\ \hline
Platelets                  & Platelets in   the blood                                    & kiloplatelets/mL & {[}25.01, ...,   850.00{]} & 60                & 0                                                                           & 454000    & 1.1                                                              & 131                                                          & 1   & 1       & 10   & 1           \\ \hline
Serum   creatinine         & Level of   creatinine in the blood                          & mg/dL            & {[}0.50, ...,   9.40{]}    &                   &                                                                             &           &                                                                  &                                                              &     &         &      &             \\ \hline
Serum sodium               & Level of   sodium in the blood                              & mEq/L            & {[}114, ...,   148{]}      &                   &                                                                             &           &                                                                  &                                                              &     &         &      &             \\ \hline
Smoking                    & If the patient   smokes                                     & Boolean          & 0, 1                       &                   &                                                                             &           &                                                                  &                                                              &     &         &      &             \\ \hline
Time                       & Follow-up   period                                          & Days             & {[}4,...,285{]}            &                   &                                                                             &           &                                                                  &                                                              &     &         &      &             \\ \hline
(target) death   event     & If the patient   died during the follow-up period           & Boolean          & 0, 1                       &                   &                                                                             &           &                                                                  &                                                              &     &         &      &             \\ \hline
\end{tabular}%
}
\end{table}

.

\section{Proposed Methodology}
\begin{figure}[h!]
    \centering
    \includegraphics[width=0.5\textwidth,height=4cm]{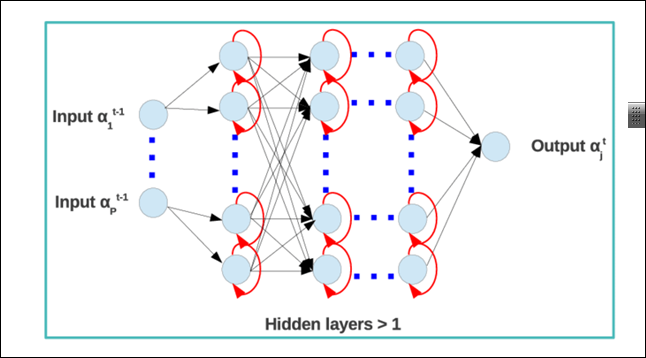}
    \caption{LSTM deep learning architecture}
    \label{f8}
    \end{figure}

Rather than using traditional SVM(Support Vector Machine) \cite{c21} model we have proposed LSTM based deep leariong model which is normally is used to decide what information we’re going to throw away from the cell state. This decision is made by a sigmoid layer called the forget gate layer. It looks at $h_{t-1}$ and $x_{t}$, and outputs a number between 0 and 1 for each number in the cell state $C_{t-1}$. A 1 represents completely keep this while a 0 represents completely get rid of this\cite{c22}.
Figure \ref{f8} represent the structure of a LSTM deep neural network  to predict the next word based on all the previous ones. In such a problem, the cell state might include the gender of the present subject, so that the correct pronouns can be used. When we see a new subject, we want to forget the gender of the old subject \cite{c23}.
In this experiment we analyze and predict the trend of heart failure utilizing the data set given in. We used LSTM deep learning model, machine learning technique for that purpose in this work. The structure of a LSTM deep neural network can be shown in figure \ref{f8}.This type of network has cyclic connections, which make the network a powerful method to model temporal data since it has an internal memory system to deal with temporal sequence inputs \cite{c24}. The LSTM has a special memory block in the hidden layer of the recurrent neural network.The architecture of a single memory block is shown in figure 9.The input gate of each memory block controls the information transmitting from the input activation's into the cell and the output gate controls the information transmitting from the memory cell activation's into other nodes. We implemented the method with Python programming language in open source Jupyter Notebook \cite{c25}.
\begin{equation}\label{masud6}
 f_{t}=\sigma (W_{f}.[h_{t-1},x_{t}] + b_{f})
\end{equation}
The next step is to decide what new information we’re going to store in the cell state. This has two parts. First, a sigmoid layer called the “input gate layer” decides which values we’ll update. Next, a tanh layer creates a vector of new candidate values,$C_{t}$, that could be added to the state. In the next step, we’ll combine these two to create an update to the state. In the language model, we’d want to add the gender of the new subject to the cell state, to replace the old one we’re forgetting \cite{c26}.

\begin{equation}\label{masud1}
 i_{t}=\sigma (W_{i}.[h_{t-1},x_{t}] + b_{i})
\end{equation}

\begin{equation}\label{masud2}
 C_{t}= tanh (W_{c}.[h_{t-1},x_{t}] + b_{c})
\end{equation}

It’s now time to update the old cell state,$C_{t-1}$, into the new cell state $C_{t}$. The previous steps already decided what to do, it just need to actually do it. Multiply the old state by $f_{t}$, forgetting the things that is decided to forget earlier \cite{c27} Then add $i_{t}$*$C_{t}$. This is the new candidate values, scaled by how much it is decided to update each state value. In the case of the language model at diagram (c) of figure 8, this is where we’d actually drop the information about the old subject’s gender and add the new information, as we decided in the previous steps\cite{c28}.

\begin{equation}\label{masud3}
 C_{t}= f_{t} *C_{t-1} + i_{t} * C_{t} 
\end{equation}
Finally, we need to decide what we re going to output. This output will be based on our cell state, but will be a filtered version. First, we run a sigmoid layer which decides what parts of the cell state we’re going to output\cite{c29}. Then, we put the cell state through  tanh to push the values to be between \{-1\} and $1$ and multiply it by the output of the $sigmoid$ gate, so that we only output the parts we decided to. For the language model’s diagram (d) of figure $8$, since it just saw a subject, it might want to output information relevant to a verb, in case that’s what is coming next. For example, it might output whether the subject is singular or plural, so that we know what form a verb should be conjugated into if that’s what follows next \cite{c30}.

\begin{equation}\label{masud4}
 o_{t}= \sigma (W_{o} * [h_{t-1},x_{t}] + b_{0}) 
\end{equation}

\begin{equation}\label{masud5}
 h_{t}= 0_{t} * tanh(C_{t})  
 \end{equation}
 
 In this experiment we analyze and predict the trend of heart failure utilizing the data set given in \cite{c31}. We used LSTM deep learning model, machine learning technique for that purpose in this work. This type of network has cyclic connections, which make the network a powerful method to model temporal data since it has an internal memory system to deal with temporal sequence inputs \cite{c32}. The LSTM has a special memory block in the hidden layer of the recurrent neural network. The input gate of each memory block controls the information transmitting from the input activation's into the cell and the output gate controls the information transmitting from the memory cell activation's into other nodes. We implemented the method with Python programming language in open source Jupyter Notebook \cite{c33}.
 
 \section{Experimental Evaluations}
In this section the computational results of our proposed method are discussed and summarized. 
 
 \subsection {Creatinine Phosphokinase vs. Age}
\begin{figure}[h]
    \centering
    \includegraphics[width=0.3\textwidth,height=2cm]{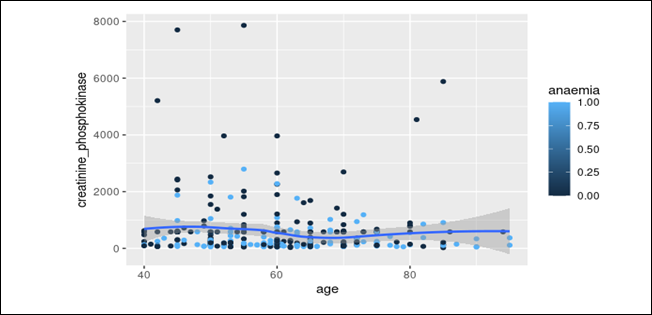}
    \caption{Creatinine Phosphokinase with respect to age}
    \label{f10}
    \end{figure}

    Figure \ref{f10} represents the plot on age vs. Creatinine Phosphokinase. From the figure, it can be see that the most of the individuals who has Creatinine Phosphokinase lie between ages 40 to 70. In the figure light blue marked dot indicate the presence of anaemia and dark blue marked dot indicate the absence of anaemia. Most interesting fact is that after age 80, most individuals have anaemia.
   
    \subsection{Diabetes vs. Age}
\begin{figure}[h]
    \centering
    \includegraphics[width=0.5\textwidth,height=4cm]{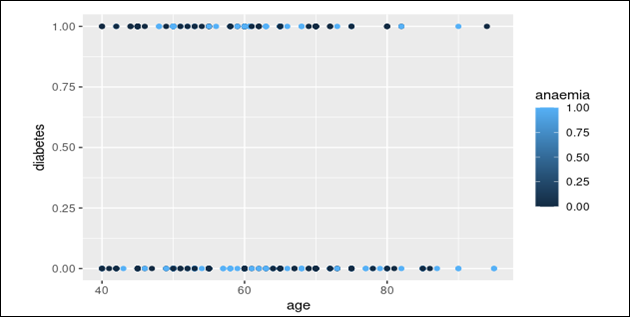}
    \caption{Diabetes according to age}
    \label{f11}
    \end{figure}

    Figure \ref{f11}
    represents a plot on age vs. Diabetes. From the figure, it can be seen that the most of the individuals has not diabetes. In the figure light blue marked dot indicate the presence of anaemia and dark blue marked dot indicate the absence of anaemia. So it conclude that diabetes less individuals have more presence of anaemia.

 \subsection {High Blood Pressure vs. Age}  
\begin{figure}[h]
    \centering
    \includegraphics[width=0.5\textwidth,height=4cm]{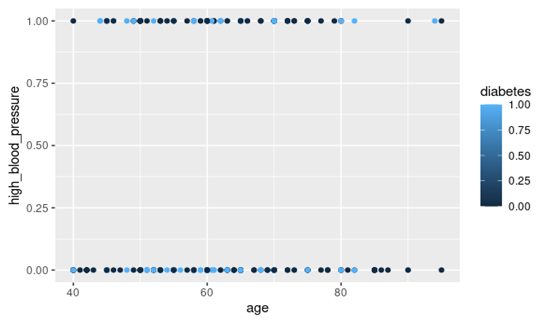}
    \caption{High blood pressure with respect to age}
    \label{f12}
    \end{figure}

    Figure \ref{f12} represents a plot on age vs. high blood pressure. From the figure, it can be seen that the most of the individuals has not high blood pressure. In the figure light blue marked dot indicate the presence of diabetes and dark blue marked dot indicate the absence of diabetes. So it say that high blood pressure less individuals have more presence of diabetes.
   
\subsection{Platelets vs. Age} 
\begin{figure}[h]
    \centering
    \includegraphics[width=0.5\textwidth,height=4cm]{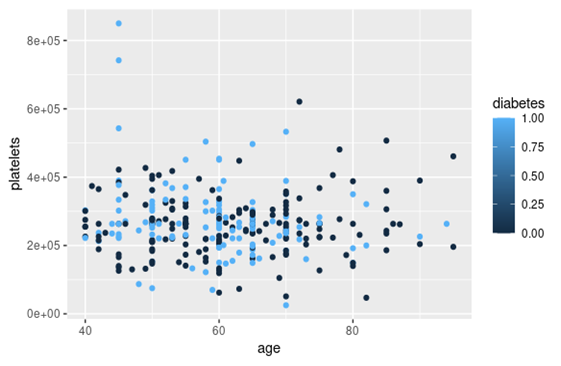}
    \caption{Platelets after age}
    \label{f13}
    \end{figure}

    Figure \ref{f13} represents a plot on age vs. Platelets. From the figure, it is found a variation of number of platelets according of the age of the individuals. In the figure light blue marked dot indicate the presence of diabetes and dark blue marked dot indicate the absence of diabetes. So it reveals that most of the individuals have presence of diabetes.
    
\subsection{Serum Sodium vs. Age} 
\begin{figure}[h]
    \centering
    \includegraphics[width=0.5\textwidth,height=4cm]{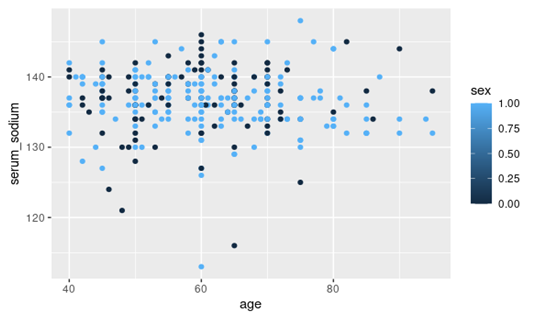}
    \caption{Serum sodium in accordance with age}
    \label{f14}
    \end{figure}

    Figure \ref{f14} represents a plot on age vs. Serum Sodium. In the figure light blue marked dot indicate the male individuals and dark blue marked dot indicate the female individuals. It indicates that most of the individuals are male in the figure. From the figure, it can be seen that highest and also lowest Serum Sodium is found in mail individuals.
    
\subsection{Total number of heart failure w.r.t age} 
\begin{figure}[h]
    \centering
    \includegraphics[width=0.5\textwidth,height=4cm]{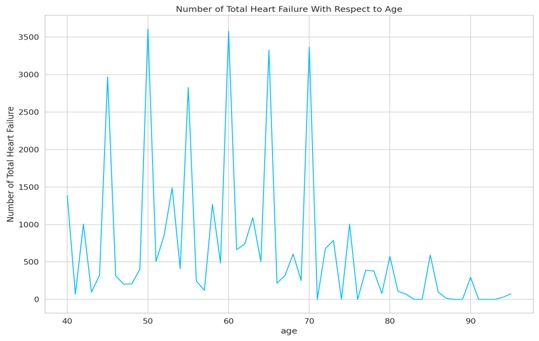}
    \caption{Total number of heart failure with respect to age}
    \label{f15}
    \end{figure}

    Total number of heart failure with respect to age is shown in figure \ref{f15}. From the figure, it can be found that highest probability of heart failure is identified around age 40 to age 60. Also, from the age 70 to higher this probability is found less.
    
\subsection{Future prediction vs. historical data} 
\begin{figure}[h]
    \centering
    \includegraphics[width=0.5\textwidth,height=4cm]{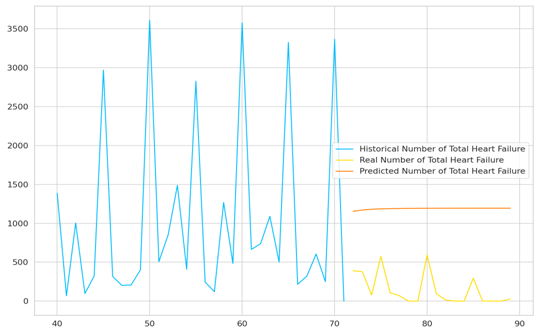}
    \caption{Future prediction and historical data}
    \label{f16}
    \end{figure}

    Figure \ref{f16} represents future prediction vs. historical data. In the figure light blue marked curve indicate the historical number of total heart failure, yellow marked curve indicate the real number of total heart failure, and red marked curve indicate the predicted number of total heart failure.
 \subsection{Future prediction  vs. historical data using all data}    
\begin{figure}[h]
    \centering
    \includegraphics[width=0.5\textwidth,height=4cm]{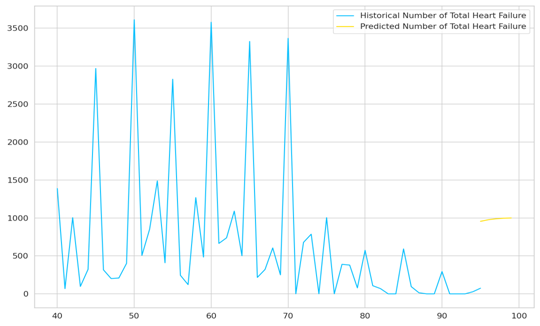}
    \caption{Future prediction and historical data}
    \label{f17}
    \end{figure}

    Figure \ref{f17} represents future prediction vs. historical data using all data. In the figure light blue marked curve indicate the historical number of total heart failure, yellow marked curve indicate the predicted number of total heart failure.    
    
\subsection{Predicted total number of heart  failure using all data}  
    
\begin{figure}[h]
    \centering
    \includegraphics[width=0.5\textwidth,height=4cm]{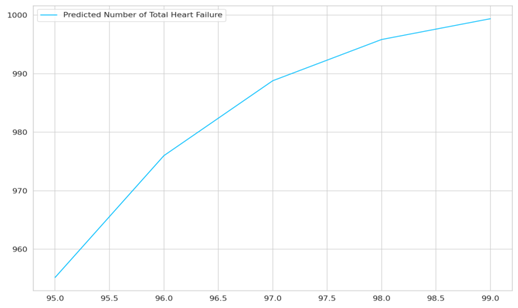}
    \caption{Future prediction and historical data}
    \label{f18}
    \end{figure}

    The predicted total number of heart failure using all data is illustrated in figure \ref{f18}. From the figure, clearly, we can see that there is an increasing trend in total number of heart failure.
    
  \subsection{Confusion matrices using part of data as training}   
    
\begin{figure}[h]
    \centering
    \includegraphics[width=0.5\textwidth,height=4cm]{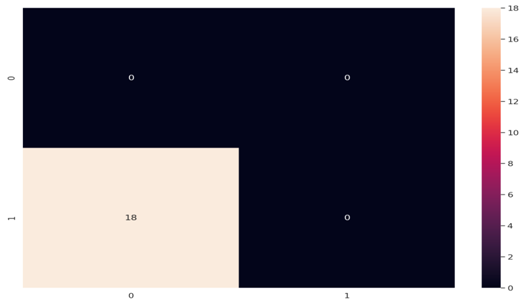}
    \caption{Confusion matrices while using part of the training data}
    \label{f19}
    \end{figure}

    Figure \ref{f19} demonstrate the confusion matrices using part of training data. From the figure, it can be noticed that the maximum value is 18 where the minimum value is 0. 
    
  \subsection{Confusion matrices using all data}   
    
\begin{figure}[h]
    \centering
    \includegraphics[width=0.5\textwidth,height=4cm]{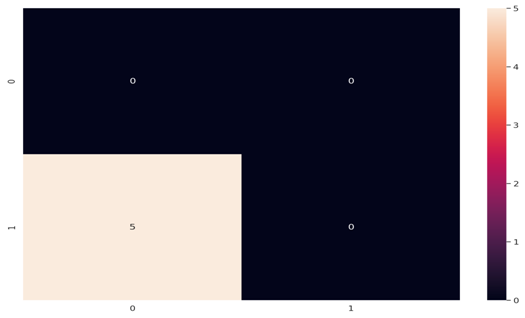}
    \caption{Confusion matrices while using all data}
    \label{f20}
    \end{figure}

    Figure \ref{f20} demonstrate the confusion matrices in the case of using all data. From the figure, it can be narrated that the maximum value is 5 where the minimum value is 0.
    
   \subsection{ Loss analysis}
  
\begin{figure}[h]
    \centering
    \includegraphics[width=0.5\textwidth,height=4cm]{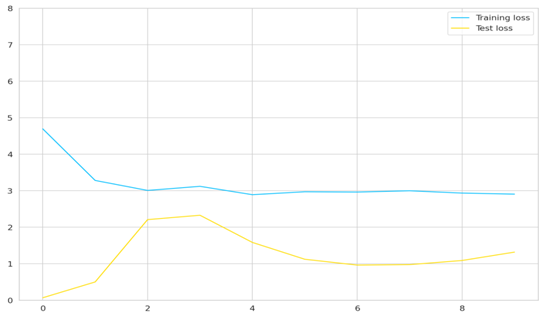}
    \caption{Training and testing loss}
    \label{f21}
    \end{figure}

    Figure \ref{f21} represents the training and testing loss of the LSTM model. Here Yellow color marked curve indicate the testing loss and light blue color marked curve indicate the training loss. From the figure, we can see that the training loss curve always reside in the higher from the test loss curve.
    
In this work, an experiment for numerical prediction of the number of the heart failure is carried out by using LSTM deep neural network model. It is found most alarming result that the predicted total number of heart failure is near to a value of 1000. Confusion matrix and loss analysis of the model is also investigated. Maximum training and testing loss is found 5 and 2, respectively

\section{Conclusion}
Heart failure (HF) is an interpretative circumstance that develops when a heart pumping system doesn’t pump enough blood circulation for the body’s requirements. It must a happened when enough blood isn't properly filled up heard. It will turn out to happen when the heart is too weak to pump blood circulation properly. On the contrary, this heart failure condition is a very serious bad patch for the human body and most of the causes are death. Over and above 6 million adults have heart failure only in the United States, in conformity with the Centers for Disease Control and Prevention. This health topic most probably focuses on heart failure in individual adults person but Children can also have heart failure. Suddenly Heart failure can happen (the acute kind) or at the time the heart stops working(the chronic kind). Left-sided and right-sided, both sides of the heart can be affected.Most of the time Left side and right side heart failure convey dissimilar causes. occasionally, once more medical condition that harm the heart and obstacles to the body's system. For  This includes coronary . We have shown an LSTM-based deep neural network model framework to predict the extensive trend of HF. In predicting HF diagnosis, LSTM models showed superior performance when compared to common methods like KNN, logistic regression, SVM, and MLP. By examining the outcomes, we have defined the significance of concerning the sequential nature of medical records. Using LSTM neural network model, we analyze a data set that consists of 13 features. This features explicitly the most responsible characteristics in the influence of causing HF. We consider a data set that has
a medical record of 299 heart failure patients collected at the Faisalabad Institute of Cardiology and the Allied Hospital in Faisalabad. The most alarming result found from our thesis work is that the predicted total number of heart failure meets a value of 1000. We also investigated the confusion matrix and the loss analysis for our LSTM model. Maximum training and testing loss is achieved at 5 and 2, respectively.

\bibliographystyle{unsrt}  


\end{document}